\pdfoutput=1

\documentclass[11pt,dvipsnames]{article}

\usepackage{acl}

\usepackage{times}
\usepackage{latexsym}

\usepackage[T1]{fontenc}

\usepackage[utf8]{inputenc}

\usepackage{microtype}

\usepackage{xcolor}
\usepackage{amsmath}

\usepackage{adjustbox} 
\usepackage{graphicx} 
\usepackage{booktabs} 
\usepackage{multirow}
\usepackage{amssymb} 
\usepackage{pifont} 
\newcommand{\cmark}{\ding{51}}%
\newcommand{\xmark}{\ding{55}}%

\newcommand{\cpm}{\mathbin{\smash{%
\raisebox{0.35ex}{%
            $\underset{\raisebox{0.5ex}{$\smash -$}}{\smash+}$%
            }%
        }%
    }%
}

\title{Pedagogical Word Recommendation: A novel task and dataset on personalized vocabulary acquisition for L2 learners}

\author{Jamin Shin \\
  Riiid AI Research / Seoul, Korea \\
  \texttt{jayshin.nlp@gmail.com} \\\And
  Juneyoung Park \\
  Riiid AI Research / Seoul, Korea \\
  \texttt{juneyoung.park@riiid.co} \\}

\begin{document}
\maketitle
\begin{abstract}
When learning a second language (L2), one of the most important but tedious components that often demoralizes students with its ineffectiveness and inefficiency is vocabulary acquisition, or more simply put, \textit{memorizing words}.
In light of such, a personalized and educational vocabulary recommendation system that traces a learner's \textit{vocabulary knowledge state} would have an immense learning impact as it could resolve both issues.
Therefore, in this paper, we propose and release data for a novel task called \textbf{Pedagogical Word Recommendation} (PWR). The main goal of PWR is to predict whether a given learner knows a given word based on other words the learner has already seen.
To elaborate, we collect this data via an Intelligent Tutoring System (ITS) called Santa that is serviced to $\sim$1M L2 learners who study for the standardized English exam, TOEIC. As a feature of this ITS, students can directly \textit{indicate words they do not know} from the questions they solved to create wordbooks. 
Finally, we report the evaluation results of a Neural Collaborative Filtering approach along with an exploratory data analysis and discuss the impact and efficacy of this dataset as a baseline for future studies on this task.
\end{abstract}

\section{Introduction}
Memorizing words is undoubtedly one of the most tiresome parts of studying a second language (L2). As discouraging as it is, vocabulary acquisition also plays a crucial role in L2 learning which makes it inevitable for students to struggle with new words every day. Evidently, this problem was in the spotlight in the field of second language acquisition for a long time~\cite{laufer1999vocabulary,schmitt2001developing, nation2006large,nakata2011computer}. 

Recently, there has also been an increasing focus on this topic from the NLP community ~\cite{ehara2010personalized,ehara2012mining,ehara2014formalizing,ehara2018building,settles2016trainable,settles2018second}. Among them,~\citet{ehara2010personalized,ehara2012mining,ehara2014formalizing,ehara2018building} formalizes this task as Vocabulary Prediction and provide datasets for it. However,~\citet{ehara2012mining} also argues that a) tracking what words learners \textit{already know} and b) determining the words learners \textit{should know}~\cite{nation2006large} are two different things that have to be modeled independently.

On the other hand, we believe that these are actually two sides of the \textit{same coin}, which is to recommend words that learners should know based on what they already know. This is precisely what our proposed task -- Pedagogical Word Recommendation (PWR), is about.
Furthermore, many works have already shown that vocabulary prediction can be a useful \textit{support} application for language learning such as TOEIC score prediction~\cite{ehara2018building}, adaptive word difficulty~\cite{ehara2012mining}, or lexical simplification
~\cite{yeung2018personalized,lee2018personalizing}. On top of this, PWR can become a core module of an ITS -- such as personalized flashcard generation.


\begin{table*}[ht!]
\centering
\begin{adjustbox}{max width=\textwidth}
\begin{tabular}{@{}cccccccccc@{}}
\toprule
Dataset                                                            & Users & Words & Data Size & Unobserved\textbf{$^\dag$} (\%) & Context Source & Dataset Purpose & Self-reported & Real-service & Timestamp \\ \midrule
\citet{ehara2010personalized,ehara2012mining}                                                 & 15    & 12k   & 180k      & 0                 & \begin{tabular}[c]{@{}c@{}}Vocabulary knowledge survey \\ (none)\end{tabular}                                         & \begin{tabular}[c]{@{}c@{}}To model user's knowledge \\ on individual words\end{tabular}                           & \cmark         & \xmark        & \xmark     \\
\midrule
\citet{ehara2018building}                                                         & 100   & 100   & 10k       & 0                 & Vocabulary exam (none)                                                                                                & \begin{tabular}[c]{@{}c@{}}To create a reliable vocabulary test \\ to measure L2 learners' vocabulary\end{tabular} & \xmark         & \xmark        & \xmark     \\
\midrule
\begin{tabular}[c]{@{}c@{}}Duolingo HLR (\textit{English} $^\ddag$)\\
\cite{settles2016trainable} \end{tabular} & 43.8k & 3k    & 5M        & 96.2              & Sentences (short)                                                                                                     & To model user's forgetting behavior                                                                                & \xmark         & \cmark        & \cmark     \\
\midrule
\begin{tabular}[c]{@{}c@{}}Duolingo SLA (\textit{English} \textbf{$^\ddag$})\\
\cite{settles2018second}\end{tabular} & 2.6k  & 2.4k  & 3.4M      & 44.8              & Sentences (short)                                                                                                     & \begin{tabular}[c]{@{}c@{}}To model where translation \\ mistakes often occur\end{tabular}                         & \xmark         & \cmark        & \cmark     \\ 
\midrule
\midrule
\textbf{PWR (ours)}                                                         & 19.7k & 5.8k  & 36.1M     & 68.4              & \begin{tabular}[c]{@{}c@{}}TOEIC Reading \& Listening\\ Comprehension  questions \\ (both short \& long)\end{tabular} & \begin{tabular}[c]{@{}c@{}}To recommend words \\ for contextual understanding\end{tabular}                         & \cmark         & \cmark        & \cmark    
\end{tabular}
\end{adjustbox}
\caption{Comparison among vocabulary prediction datasets. Our proposed dataset is more suitable for a recommendation system than others as it not only is the largest in size, but also has more desirable traits (words come from various context, self-reported, real-service, timestamps). \textbf{\dag}: \textit{Unobserved} here indicates the ratio of unobserved user-word pairs. \textbf{\ddag}: Both Duolingo datasets are \textit{multilingual} ones so we calculate statistics only from their English data for fair comparison.}
\label{tab:dataset_comparison}
\end{table*}

\paragraph{Our Contributions.} 
The contributions of this paper can be summarized into three points:
\begin{itemize}
    \item We formalize a novel task Pedagogical Word Recommendation that provides a recommendation system perspective to vocabulary prediction.
    \item Based on this formulation, we provide a novel dataset that is both large-scale ($\sim$36M) and self-reported.\footnote{We will release full dataset at \url{https://github.com/jshin49/pwr}.}
    \item We provide detailed statistics and analysis of the dataset, and evaluate standard Collaborative Filtering models to show the efficacy of our work as a benchmark for future studies in this task.
\end{itemize}




\section{Related Work}

\citet{ehara2010personalized} and ~\citet{ehara2012mining} curated a vocabulary prediction dataset by surveying 15 English L2 learners on their knowledge of 11,999 individual words in order to model a user's vocabulary knowledge along with learner-specific word difficulty. 
Using this dataset,~\citet{ehara2014formalizing} investigated a more practical scenario of sampling a small subset of words for vocabulary prediction. 
Moreover,~\citet{ehara2018building} provides a dataset on vocabulary knowledge of 100 Japanese ESL learners on 100 words obtained from vocabulary exams.
However, our work differs in several important aspects from the above works as PWR data 1) is much larger in scale, 2) comes from real-service data with words reported from TOEIC questions, and 3) has timestamps (for temporal models).

To continue, real-data from the popular L2 learning ITS Duolingo has also been released by~\cite{settles2016trainable,settles2018second}. 
The former~\cite{settles2016trainable} releases a large dataset with word-recall related features in order to model forgetting behaviors of learners.
Meanwhile, the latter~\cite{settles2018second} proposes a new task that aims to predict where the learners often make mistakes during translation and is highly associated with grammatical error detection/correction.
The main differences with our work are that PWR data 1) is self-reported which is a crucial aspect for recommendation, 2) has much longer context length for the words, and 3) is more focused on vocabulary recommendation (or prediction).

Furthermore, these works are highly related to Knowledge Tracing~\cite{corbett1994knowledge}, which is a heavily studied task~\cite{feng2009addressing,lindsey2014automatic,choi2020ednet} that aims to predict the learner's future performance on selected \textit{knowledge components} (e.g. questions or concepts) given the learner's historical data. 
~\citet{ehara2012mining,ehara2014formalizing,settles2016trainable,settles2018second} can be viewed as an instance of Knowledge Tracing, 
but these works are not positioned for the task we propose. For example, these works may not apply well to more realistic language learning scenarios like reading comprehension. As shown in Table~\ref{tab:dataset_comparison}, PWR originates from such situations making it more appropriate.
We believe that PWR combines the advantages of ITS~\cite{settles2016trainable,settles2018second} with the formalizations from~\citet{ehara2012mining,ehara2018building}.

\section{Task \& Dataset Description}
\subsection{Task Definition}
\begin{figure}[t]
\includegraphics[width=\columnwidth]{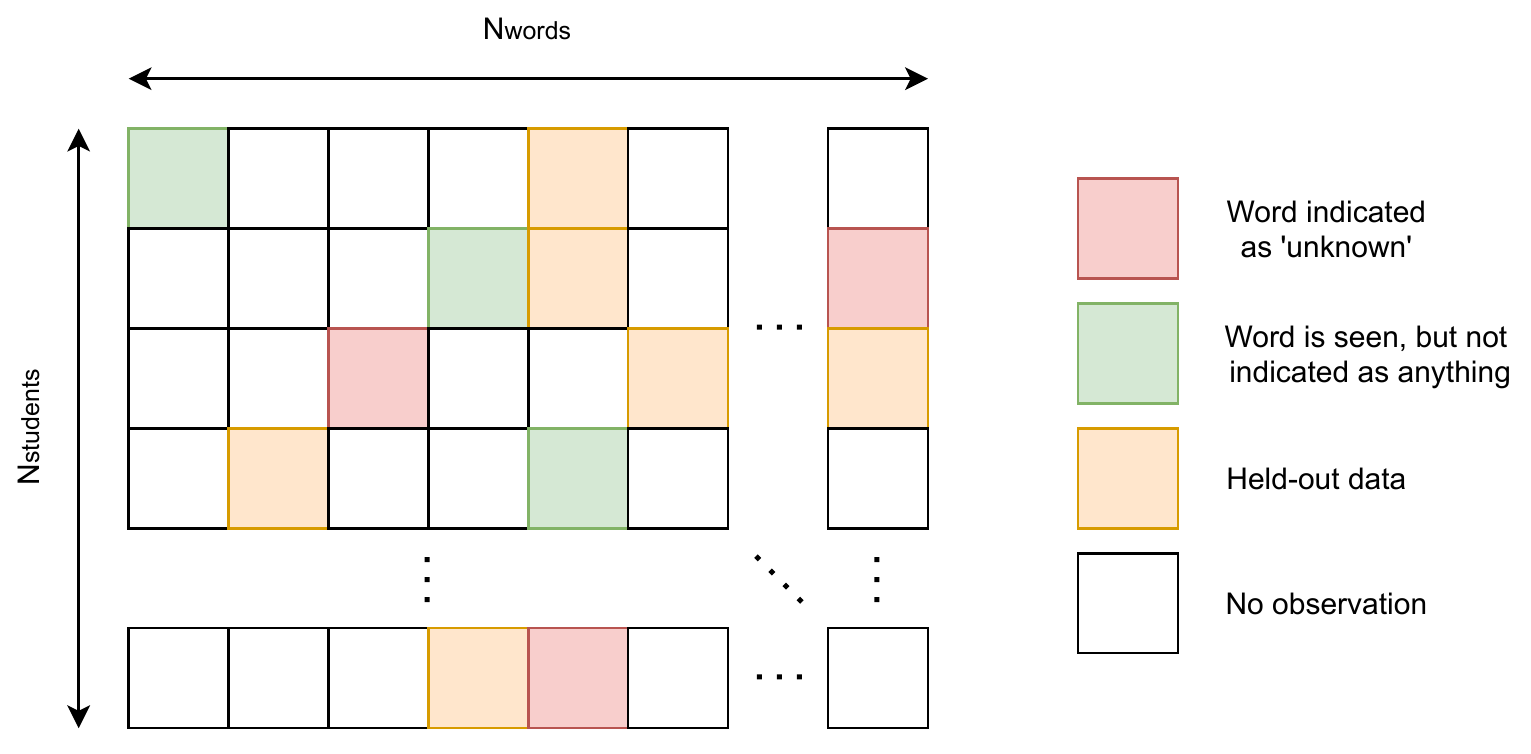}
\caption{Collaborative Filtering setting: Given the \textcolor{Red}{seen-and-unknown words} and \textcolor{LimeGreen}{seen-but-known words} as training set, a model has to predict the values of the \textcolor{Apricot}{held out words} as a test set.}
\label{fig:cf}
\end{figure}

The task definition is illustrated in Figure~\ref{fig:cf} and is similar to that of~\citet{ehara2010personalized,ehara2012mining}
\footnote{Note that in our dataset, positives indicate words that users do not know, as opposed to~\citet{ehara2010personalized,ehara2012mining}.}: 
given a user \textit{i} (or learner) and a word \textit{j}, predict whether he or she knows this word (binary classification). 
Given such, we can see that our proposed task follows the Collaborative Filtering~\cite{su2009survey} setting. Our task can be considered similar to classic recommendation system settings, but just binary~\cite{verstrepen2017collaborative} instead of ratings. \textit{The underlying assumption is that similar users ``do not know" similar words.} 
This setting is analogous to that of NeurIPS Education Challenge Task 1~\cite{wang2020diagnostic}, if we interchange ``questions" with ``words". 

\subsection{Dataset Description}
We collect our data mainly from an Intelligent Tutoring System (ITS) called Santa\footnote{\url{https://www.riiid.co/}} that is live serviced to $\sim$1M English L2 learners who are preparing for TOEIC. Users of this web \& mobile app are able to study for (mock) TOEIC Reading \& Listening Comprehension questions. For each question they have solved, users are able to review the full passages (or transcripts for Listening questions) and mark the words they do not know, and append them to their wordbooks. 
Basic summary statistics are shown in the top row group of Table~\ref{tab:statistics}.

From the joint distribution plot in Figure~\ref{fig:kde}, we can see an interesting concentration on 0\% and 100\% accuracy that correspond to the very low number of words per user. We can also see that the "high" number of unknown words is mostly associated with accuracy between 50-70\%. These observations can be explained by actual learning behaviors of L2 learners: 1) students on both extremes either know too much or too few words to create a wordbook, 2) while students in the middle are more active in studying vocabulary.

Meanwhile, we also look at the top-20 words that users know and do not know in Table~\ref{tab:top_words}. From this table, we can see that the words that are marked words are indeed difficult ones for beginners, and the words that are popularly known are seemingly easier ones.
We provide details on how we prepare this dataset along with additional dataset analysis in Appendix B and C.

\begin{table}[t]
\centering
\begin{adjustbox}{max width=\columnwidth}
\begin{tabular}{@{}ccccc@{}}
\toprule
Name                  & Train         & Dev        & Test      & Total             \\ \midrule
\# Unique users       & 19734         & 19731      & 19733     & 19734             \\
\# Unique words       & 5718          & 5277       & 5286      & 5776              \\
\# Cold-start words   & -             & 30         & 33        & 58               \\
\# Positives (+)      & 411251        & 55388      & 63180     & 529819            \\
\# Negatives (-)      & 35460094      & 63225      & 63435     & 35586754          \\ \midrule
\# Words per user (+) & \multicolumn{3}{c}{\multirow{2}{*}{-}} & 26.85 / 55.68     \\
\# Users per word (+) & \multicolumn{3}{c}{}                   & 91.74 / 153.12    \\
\# Words per user (-) & \multicolumn{3}{c}{\multirow{2}{*}{-}} & 1803.41 / 1327.71 \\
\# Users per word (-) & \multicolumn{3}{c}{}                   & 7498.26 / 3994.35 \\ \bottomrule
\end{tabular}
\end{adjustbox}
\caption{Summary statistics (top rows) and distributional statistics (bottom rows). For the bottom rows, the left is the \textit{mean} and the right is \textit{standard deviation} of counts. There is high imbalance between positive and negative labels in the training set, but the dev/test sets are balanced for proper evaluation.
}
\label{tab:statistics}
\end{table}


\begin{table}[t]
\centering
\begin{adjustbox}{max width=\columnwidth}
\begin{tabular}{@{}cc@{}}
\toprule
Type                                                                            & Top-20 Words                                                                                                                                                                                                                                                            \\ \midrule
\begin{tabular}[c]{@{}c@{}}Top words\\ unknown to users\end{tabular} & \begin{tabular}[c]{@{}c@{}}patron, inquiry, imperative, implement, complimentary, tenant, \\ representative, property, banquet, breakthrough, assume, \\ acquisition, retract, volatile, endorsement, deem, assortment, \\ preliminary, feasible, executive\end{tabular} \\ \midrule
\begin{tabular}[c]{@{}c@{}}Top words\\ known to users\end{tabular}   & \begin{tabular}[c]{@{}c@{}}available, receive, complete, while, issue, include, contact,\\ therefore, attend, free, book, additional, upcoming, current\\ request, purchase, submit, following, appreciate, several\end{tabular}                                        \\ \bottomrule
\end{tabular}
\end{adjustbox}
\caption{Words that had most number of users. Top row is words that the largest number of users found difficult (\textit{\# Users per word ($+$)}), and bottom row is words that most users knew (\textit{\# Users per word ($-$)}).}
\label{tab:top_words}
\end{table}

\section{Evaluation}
\label{sec:evaluation}
\subsection{Collaborative Filtering Models}
We evaluate our dataset on two well-known model-based Collaborative Filtering methods: Matrix Factorization (\texttt{MF})~\cite{mfcollaborative} and Neural Collaborative Filtering (\texttt{NCF})~\cite{neuralcollaborative}. 
The inputs of these models are given as user $u_i$, word $v_j$, and label $y_{ij}$. To get the output prediction $\hat{y}_{ij}$, collaborative filtering models follow a general framework that is composed of two components: 1) embedding models for users ($\mathbf{U}$) and words ($\mathbf{V}$), and an arbitrary similarity scoring function ($f$) that takes in user ($\mathbf{u}$) and word ($\mathbf{v}$) embedding vectors.
\begin{gather}
    \mathbf{u} = \mathbf{U}(u_i) \quad \mathbf{v} = \mathbf{V}(v_j) \label{eq:1} \\
    \hat{y}_{ij} = \sigma\big( f(\mathbf{u}, \mathbf{v}) \big)
    \label{eq:2}
\end{gather}
, where $\sigma(\cdot)$ is the Sigmoid function. The training objective is to minimize the binary cross entropy loss between $y_{ij}$ and $\hat{y}_{ij}$.
For the former model \texttt{MF}, $f$ is simply a \textit{dot product} operation; hence, Equation~\ref{eq:2} becomes $\sigma(\mathbf{u} \cdot \mathbf{v})$. On the other hand, for the latter model \texttt{NCF}, $f$ is a multi-layer perceptron network that takes $[\mathbf{u}; \mathbf{v}]$ as input (concatenation).
In addition, 
we initialize $\mathbf{V}$ with pre-trained \texttt{word2vec} embeddings~\cite{mikolov2013distributed}.

\begin{figure}[t]
\includegraphics[width=\columnwidth]{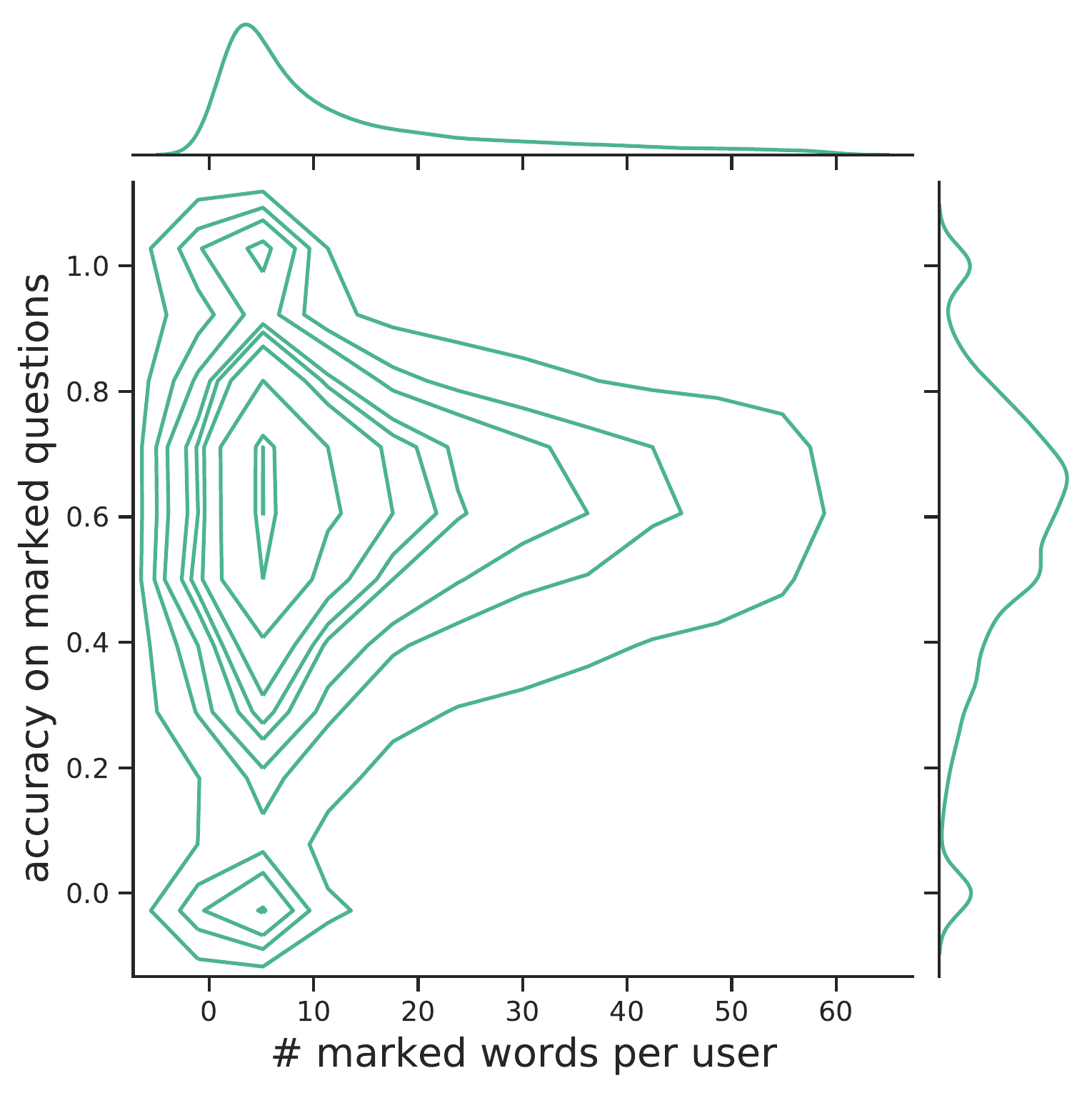}
\caption{Joint plot on the marginal distributions. The y-axis accuracy is calculated only on the questions that the marked words come from. Other versions (hexbin \& scatter plots) are available in Appendix C.}
\label{fig:kde}
\end{figure}

\subsection{Experiment Setup}
In this section, we describe how we set up the experiments in order to evaluate the usability of this PWR dataset. We use accuracy, precision, recall, and f1-score as metrics for evaluation. We experiment using \texttt{word2vec} as initialization and also experiment with different number of layers for \texttt{NCF}.
Appendix A describes in detail the hyperparameters and other training details.

\subsection{Results}
Table~\ref{tab:test_results} summarizes the experiment results on the test set. In general, we can see that adding \texttt{word2vec} to both \texttt{MF} and \texttt{NCF}$_{*-\text{layer}}$ helps the performance -- whether the effect is significant or minor. Interestingly, the effect of \texttt{word2vec} is much stronger on \texttt{MF} which can be attributed to the smaller model capacity. We can also see that the models are overall making balanced predictions as accuracy and F1-score are similar. Finally, while the comparison between \texttt{NCF}$_{*-\text{layer}}$ is inconclusive, they certainly outperform \texttt{MF}. From these results, we show that PWR can serve as a challenging benchmark dataset. 
Similar evaluation results on the dev split are shown in Appendix A.

\begin{table}[t]
\centering
\begin{adjustbox}{max width=\columnwidth}
\begin{tabular}{@{}ccccc@{}}
\toprule
\textbf{Test-set}             & Accuracy                    & Precision                 & Recall                     & F1-score                   \\ \midrule
$\texttt{MF}$                 & $0.56 \cpm 0.002$           & $0.61 \cpm 0.004$         & $0.36 \cpm 0.006$          & $0.45 \cpm 0.004$          \\
$+ \texttt{word2vec}$         & $0.6 \cpm 0.001$            & $0.67 \cpm 0.003$         & $0.4 \cpm 0.003$           & $0.5 \cpm 0.002$           \\ \midrule
$\texttt{NCF}_\text{1-layer}$ & 0.696 $\cpm$ 0.001 & $0.729 \cpm 0.005$        & $0.62 \cpm 0.01$           & $0.67 \cpm 0.004$          \\
$+ \texttt{word2vec}$         & $0.686 \cpm 0.003$          & $0.71 \cpm 0.004$         & $0.63 \cpm 0.01$           & $0.67 \cpm 0.007$          \\ \midrule
$\texttt{NCF}_\text{2-layer}$ & \textbf{0.698 $\cpm$ 0.003}          & $0.733 \cpm 0.007$        & $0.62 \cpm 0.01$          & $0.67 \cpm 0.003$          \\
$+ \texttt{word2vec}$         & $0.692 \cpm 0.003$          & $0.71 \cpm 0.003$         & \textbf{0.64 $\cpm$ 0.01} & \textbf{0.68 $\cpm$ 0.004} \\ \midrule
$\texttt{NCF}_\text{3-layer}$ & $0.69 \cpm 0.01$            & \textbf{0.74 $\cpm$ 0.01} & $0.57 \cpm 0.04$           & $0.64 \cpm 0.03$           \\
$+ \texttt{word2vec}$         & $0.69 \cpm 0.003$            & $0.72 \cpm 0.005$         & \textbf{0.64 $\cpm$ 0.01}  & \textbf{0.68 $\cpm$ 0.004} \\ \bottomrule
\end{tabular}
\end{adjustbox}
\caption{Experiment results on test set. Each model is run 5 times with different seeds; hence, each result entry is in the form of mean $\cpm$ standard deviation.}
\label{tab:test_results}
\end{table}

\section{Discussion on Efficacy}
In this section, we discuss the implications of PWR to the research community. 
First of all, vocabulary prediction itself can be a core component to other NLP tasks like lexical simplification~\cite{yeung2018personalized,lee2018personalizing} or automated essay scoring~\cite{taghipour2016neural}.~\citet{ehara2018building} has already shown that TOEIC score prediction has a high correlation with vocabulary predictions. 
Similarly, we can try to improve neural essay scoring~\cite{taghipour2016neural} with our dataset. 
Secondly, we believe that our work can fundamentally improve computer-assisted language learning through efficient and effective personalized vocabulary acquisition. 
For example, as mentioned before, an automated flashcard generator would be able to significantly increase both efficiency and effectiveness of memorizing words. 
Finally, PWR can more realistically extend Knowledge Tracing to vocabulary-level. 
There were also a few works that incorporate question text information for Knowledge Tracing~\cite{liu2019ekt,pandey2020rkt,tong2020hgkt}, but these works mostly focused on utilizing word-level embeddings for obtaining question embeddings. 
On the other hand, applying Deep Knowledge Tracing models~\cite{piech2015deep,zhang2017dynamic,ghosh2020context} to PWR will allow us to have a better understanding of user knowledge states.


\section{Conclusion}
In this paper, we focus on solving the inefficiency and ineffectiveness of vocabulary acquisition during second language (L2) learning. For this purpose, we formalized and introduced a novel task called Pedagogical Word Recommendation that adds a recommendation system perspective to the conventional vocabulary prediction task. We provide a novel dataset that is large-scale ($\sim$36M) for this task and describe how it was created along with detailed statistics and analysis. Finally, we show the efficacy of this dataset not only as a benchmark but for a test-bed of bridging the gap between NLP and education.


\bibliography{anthology,custom}

\begin{thebibliography}{29}
\expandafter\ifx\csname natexlab\endcsname\relax\def\natexlab#1{#1}\fi

\bibitem[{Choi et~al.(2020)Choi, Lee, Shin, Cho, Park, Lee, Baek, Bae, Kim, and
  Heo}]{choi2020ednet}
Youngduck Choi, Youngnam Lee, Dongmin Shin, Junghyun Cho, Seoyon Park, Seewoo
  Lee, Jineon Baek, Chan Bae, Byungsoo Kim, and Jaewe Heo. 2020.
\newblock Ednet: A large-scale hierarchical dataset in education.
\newblock In \emph{International Conference on Artificial Intelligence in
  Education}, pages 69--73. Springer.

\bibitem[{Corbett and Anderson(1994)}]{corbett1994knowledge}
Albert~T Corbett and John~R Anderson. 1994.
\newblock Knowledge tracing: Modeling the acquisition of procedural knowledge.
\newblock \emph{User modeling and user-adapted interaction}, 4(4):253--278.

\bibitem[{Ehara(2018)}]{ehara2018building}
Yo~Ehara. 2018.
\newblock Building an english vocabulary knowledge dataset of japanese
  english-as-a-second-language learners using crowdsourcing.
\newblock In \emph{Proceedings of the Eleventh International Conference on
  Language Resources and Evaluation (LREC 2018)}.

\bibitem[{Ehara et~al.(2014)Ehara, Miyao, Oiwa, Sato, and
  Nakagawa}]{ehara2014formalizing}
Yo~Ehara, Yusuke Miyao, Hidekazu Oiwa, Issei Sato, and Hiroshi Nakagawa. 2014.
\newblock Formalizing word sampling for vocabulary prediction as graph-based
  active learning.
\newblock In \emph{Proceedings of the 2014 Conference on Empirical Methods in
  Natural Language Processing (EMNLP)}, pages 1374--1384.

\bibitem[{Ehara et~al.(2012)Ehara, Sato, Oiwa, and Nakagawa}]{ehara2012mining}
Yo~Ehara, Issei Sato, Hidekazu Oiwa, and Hiroshi Nakagawa. 2012.
\newblock Mining words in the minds of second language learners:
  learner-specific word difficulty.
\newblock In \emph{Proceedings of COLING 2012}, pages 799--814.

\bibitem[{Ehara et~al.(2010)Ehara, Shimizu, Ninomiya, and
  Nakagawa}]{ehara2010personalized}
Yo~Ehara, Nobuyuki Shimizu, Takashi Ninomiya, and Hiroshi Nakagawa. 2010.
\newblock Personalized reading support for second-language web documents by
  collective intelligence.
\newblock In \emph{Proceedings of the 15th international conference on
  Intelligent user interfaces}, pages 51--60.

\bibitem[{Feng et~al.(2009)Feng, Heffernan, and Koedinger}]{feng2009addressing}
Mingyu Feng, Neil Heffernan, and Kenneth Koedinger. 2009.
\newblock Addressing the assessment challenge with an online system that tutors
  as it assesses.
\newblock \emph{User modeling and user-adapted interaction}, 19(3):243--266.

\bibitem[{Ghosh et~al.(2020)Ghosh, Heffernan, and Lan}]{ghosh2020context}
Aritra Ghosh, Neil Heffernan, and Andrew~S Lan. 2020.
\newblock Context-aware attentive knowledge tracing.
\newblock In \emph{Proceedings of the 26th ACM SIGKDD International Conference
  on Knowledge Discovery \& Data Mining}, pages 2330--2339.

\bibitem[{He et~al.(2017)He, Liao, Zhang, Nie, Hu, and
  Chua}]{neuralcollaborative}
Xiangnan He, Lizi Liao, Hanwang Zhang, Liqiang Nie, Xia Hu, and Tat-Seng Chua.
  2017.
\newblock \href {https://doi.org/10.1145/3038912.3052569} {Neural collaborative
  filtering}.
\newblock In \emph{Proceedings of the 26th International Conference on World
  Wide Web}, WWW '17, page 173–182, Republic and Canton of Geneva, CHE.
  International World Wide Web Conferences Steering Committee.

\bibitem[{Laufer and Nation(1999)}]{laufer1999vocabulary}
Batia Laufer and Paul Nation. 1999.
\newblock A vocabulary-size test of controlled productive ability.
\newblock \emph{Language testing}, 16(1):33--51.

\bibitem[{Lee and Yeung(2018)}]{lee2018personalizing}
John~SY Lee and Chak~Yan Yeung. 2018.
\newblock Personalizing lexical simplification.
\newblock In \emph{Proceedings of the 27th International Conference on
  Computational Linguistics}, pages 224--232.

\bibitem[{Lindsey et~al.(2014)Lindsey, Khajah, and
  Mozer}]{lindsey2014automatic}
Robert~V Lindsey, Mohammad Khajah, and Michael~C Mozer. 2014.
\newblock Automatic discovery of cognitive skills to improve the prediction of
  student learning.
\newblock In \emph{Advances in neural information processing systems}, pages
  1386--1394. Citeseer.

\bibitem[{Liu et~al.(2019)Liu, Huang, Yin, Chen, Xiong, Su, and
  Hu}]{liu2019ekt}
Qi~Liu, Zhenya Huang, Yu~Yin, Enhong Chen, Hui Xiong, Yu~Su, and Guoping Hu.
  2019.
\newblock Ekt: Exercise-aware knowledge tracing for student performance
  prediction.
\newblock \emph{IEEE Transactions on Knowledge and Data Engineering},
  33(1):100--115.

\bibitem[{Mikolov et~al.(2013)Mikolov, Sutskever, Chen, Corrado, and
  Dean}]{mikolov2013distributed}
Tomas Mikolov, Ilya Sutskever, Kai Chen, Gregory~S Corrado, and Jeffrey Dean.
  2013.
\newblock Distributed representations of words and phrases and their
  compositionality.
\newblock In \emph{NIPS}.

\bibitem[{Nakata(2011)}]{nakata2011computer}
Tatsuya Nakata. 2011.
\newblock Computer-assisted second language vocabulary learning in a
  paired-associate paradigm: A critical investigation of flashcard software.
\newblock \emph{Computer Assisted Language Learning}, 24(1):17--38.

\bibitem[{Nation(2006)}]{nation2006large}
I~Nation. 2006.
\newblock How large a vocabulary is needed for reading and listening?
\newblock \emph{Canadian modern language review}, 63(1):59--82.

\bibitem[{Pandey and Srivastava(2020)}]{pandey2020rkt}
Shalini Pandey and Jaideep Srivastava. 2020.
\newblock Rkt: Relation-aware self-attention for knowledge tracing.
\newblock In \emph{Proceedings of the 29th ACM International Conference on
  Information \& Knowledge Management}, pages 1205--1214.

\bibitem[{Piech et~al.(2015)Piech, Bassen, Huang, Ganguli, Sahami, Guibas, and
  Sohl-Dickstein}]{piech2015deep}
Chris Piech, Jonathan Bassen, Jonathan Huang, Surya Ganguli, Mehran Sahami,
  Leonidas~J Guibas, and Jascha Sohl-Dickstein. 2015.
\newblock Deep knowledge tracing.
\newblock In \emph{NIPS}.

\bibitem[{Schmitt et~al.(2001)Schmitt, Schmitt, and
  Clapham}]{schmitt2001developing}
Norbert Schmitt, Diane Schmitt, and Caroline Clapham. 2001.
\newblock Developing and exploring the behaviour of two new versions of the
  vocabulary levels test.
\newblock \emph{Language testing}, 18(1):55--88.

\bibitem[{Settles et~al.(2018)Settles, Brust, Gustafson, Hagiwara, and
  Madnani}]{settles2018second}
Burr Settles, Chris Brust, Erin Gustafson, Masato Hagiwara, and Nitin Madnani.
  2018.
\newblock Second language acquisition modeling.
\newblock In \emph{Proceedings of the thirteenth workshop on innovative use of
  NLP for building educational applications}, pages 56--65.

\bibitem[{Settles and Meeder(2016)}]{settles2016trainable}
Burr Settles and Brendan Meeder. 2016.
\newblock A trainable spaced repetition model for language learning.
\newblock In \emph{Proceedings of the 54th annual meeting of the association
  for computational linguistics (volume 1: long papers)}, pages 1848--1858.

\bibitem[{Su and Khoshgoftaar(2009)}]{su2009survey}
Xiaoyuan Su and Taghi~M Khoshgoftaar. 2009.
\newblock A survey of collaborative filtering techniques.
\newblock \emph{Advances in artificial intelligence}, 2009.

\bibitem[{Taghipour and Ng(2016)}]{taghipour2016neural}
Kaveh Taghipour and Hwee~Tou Ng. 2016.
\newblock A neural approach to automated essay scoring.
\newblock In \emph{Proceedings of the 2016 conference on empirical methods in
  natural language processing}, pages 1882--1891.

\bibitem[{Tong et~al.(2020)Tong, Zhou, and Wang}]{tong2020hgkt}
Hanshuang Tong, Yun Zhou, and Zhen Wang. 2020.
\newblock Hgkt: Introducing problem schema with hierarchical exercise graph for
  knowledge tracing.
\newblock \emph{arXiv preprint arXiv:2006.16915}.

\bibitem[{Verstrepen et~al.(2017)Verstrepen, Bhaduriy, Cule, and
  Goethals}]{verstrepen2017collaborative}
Koen Verstrepen, Kanishka Bhaduriy, Boris Cule, and Bart Goethals. 2017.
\newblock Collaborative filtering for binary, positiveonly data.
\newblock \emph{ACM SIGKDD Explorations Newsletter}, 19(1):1--21.

\bibitem[{Wang et~al.(2020)Wang, Lamb, Saveliev, Cameron, Zaykov,
  Hern{\'a}ndez-Lobato, Turner, Baraniuk, Barton, Jones
  et~al.}]{wang2020diagnostic}
Zichao Wang, Angus Lamb, Evgeny Saveliev, Pashmina Cameron, Yordan Zaykov,
  Jos{\'e}~Miguel Hern{\'a}ndez-Lobato, Richard~E Turner, Richard~G Baraniuk,
  Craig Barton, Simon~Peyton Jones, et~al. 2020.
\newblock Diagnostic questions: The neurips 2020 education challenge.
\newblock \emph{arXiv preprint arXiv:2007.12061}.

\bibitem[{Yeung and Lee(2018)}]{yeung2018personalized}
Chak~Yan Yeung and John~SY Lee. 2018.
\newblock Personalized text retrieval for learners of chinese as a foreign
  language.
\newblock In \emph{Proceedings of the 27th International Conference on
  Computational Linguistics}, pages 3448--3455.

\bibitem[{Zhang et~al.(2017)Zhang, Shi, King, and Yeung}]{zhang2017dynamic}
Jiani Zhang, Xingjian Shi, Irwin King, and Dit-Yan Yeung. 2017.
\newblock Dynamic key-value memory networks for knowledge tracing.
\newblock In \emph{Proceedings of the 26th international conference on World
  Wide Web}, pages 765--774.

\bibitem[{Zhang et~al.(2014)Zhang, Liu, Chun-Gui, Wei, and
  Huiyi-Ma}]{mfcollaborative}
Ruisheng Zhang, Qi-dong Liu, Chun-Gui, Jia-Xuan Wei, and Huiyi-Ma. 2014.
\newblock Collaborative filtering for recommender systems.
\newblock In \emph{Proceedings of the 2014 Second International Conference on
  Advanced Cloud and Big Data}, CBD '14, page 301–308, USA. IEEE Computer
  Society.

\end{thebibliography}
\bibliographystyle{acl_natbib}

\clearpage
\appendix
\section*{Ethical Impact Statement}
\begin{itemize}
    \item We fully anonymize the data including user IDs and release no personally identifiable information.
    \item We made sure that the users agreed to our usage of their interaction data with the ITS.
    \item As no actual content data can be recovered from our data, we made sure no Intellectual Property rights were violated.
    \item As we are not an academic institute, we do not have an Institutional Review Board (IRB), but the corresponding author completed the IRB education on Social Behavior Research provided by CITI.
    \item We consulted with lawyers and patent lawyers regarding any legal issues with this paper.
\end{itemize}

\section{Training Details}
We summarize the training details like hyper-parameters and such:
\begin{itemize}
    \item To cope with the label imbalance during training, we use down-sampling.
    \item We use Adam optimizer with Learning rate 5e$^{-3}$ for both \texttt{MF} and \texttt{NCF}$_*$.
    \item The dimensionality of both input embeddings is 300.
    \item We use batch size of 32768 for both training and evaluation.
    \item For the additional layers in \texttt{NCF}$_*$, we kept the hidden dimensions same.
    \item The \texttt{word2vec} embeddings we use are \texttt{crawl-300d-2M.vec.zip} given at fasttext website
    \url{https://fasttext.cc/docs/en/english-vectors.html}
\end{itemize}

Results on dev set and the number of parameters are given at Table~\ref{tab:dev_results}.

\begin{table*}[t]
\centering
\begin{adjustbox}{max width=\textwidth}
\begin{tabular}{@{}cccccc@{}}
\toprule
\textbf{Dev-set}              & Accuracy          & Precision         & Recall            & F1-score          & Parameters             \\ \midrule
$\texttt{MF}$                 & 0.58 $\cpm$ 0.002 & 0.58 $\cpm$ 0.004 & 0.37 $\cpm$ 0.005 & 0.45 $\cpm$ 0.003 & \multirow{2}{*}{7.65M} \\
$+ \texttt{word2vec}$         & 0.62 $\cpm$ 0.002 & 0.65 $\cpm$ 0.006 & 0.42 $\cpm$ 0.003 & 0.51 $\cpm$ 0.003 &                        \\ \midrule
$\texttt{NCF}_\text{1-layer}$ & 0.71 $\cpm$ 0.001 & 0.71 $\cpm$ 0.005 & 0.65 $\cpm$ 0.01  & 0.67 $\cpm$ 0.003 & \multirow{2}{*}{7.65M} \\
$+ \texttt{word2vec}$         & 0.70 $\cpm$ 0.002 & 0.69 $\cpm$ 0.004 & 0.65 $\cpm$ 0.01  & 0.68 $\cpm$ 0.005 &                        \\ \midrule
$\texttt{NCF}_\text{2-layer}$ & 0.71 $\cpm$ 0.003 & 0.71 $\cpm$ 0.007 & 0.64 $\cpm$ 0.007 & 0.68 $\cpm$ 0.002 & \multirow{2}{*}{7.83M} \\
$+ \texttt{word2vec}$         & 0.71 $\cpm$ 0.002 & 0.69 $\cpm$ 0.002 & 0.67 $\cpm$ 0.007 & 0.68 $\cpm$ 0.003 &                        \\ \midrule
$\texttt{NCF}_\text{3-layer}$ & 0.70 $\cpm$ 0.009 & 0.72 $\cpm$ 0.01  & 0.59 $\cpm$ 0.04  & 0.65 $\cpm$ 0.02  & \multirow{2}{*}{7.92M} \\
$+ \texttt{word2vec}$         & 0.71 $\cpm$ 0.002 & 0.70 $\cpm$ 0.006 & 0.66 $\cpm$ 0.01  & 0.68 $\cpm$ 0.003 &                        \\ \bottomrule
\end{tabular}
\end{adjustbox}
\caption{Experiment results on dev set. Each model is run 5 times with different seeds; hence, each result entry is in the form of mean $\cpm$ standard deviation.}
\label{tab:dev_results}
\end{table*}

\section{Details on Dataset Preparation}
We split the dataset and provide the splits along with it (will release for camera ready). The train/dev/test splits are split randomly following the ratio of 80-10-10. As the original dataset has highly unbalanced labels, we make sure that the dev and test splits maintain a balanced ratio for proper evaluation. The function \texttt{split\_train\_dev\_test} in the code (line 293) is where we do this operation.

Furthermore, we filter out the following:
\begin{itemize}
    \item Cold-start users (users with only 1 word)
    \item Cold-start words (words with only 1 user)
    \item noisy users (either bugs or test users)
    \item UNK words from \texttt{word2vec}
    \item Remove stop-words and entities (may not be perfect)
    \item words with no associated question IDs (possibly outdated questions)
\end{itemize}

Finally, we anonymize the dataset in a very simple way. As the only personal identifiable or copyright-sensitive information in the original data are user IDs and question IDs, we simply map user IDs to random integers, and we DO NOT release the accompanying question IDs.

\section{Additional Dataset Analysis}
\begin{figure*}[t]
\centering
\includegraphics[width=0.49\textwidth]{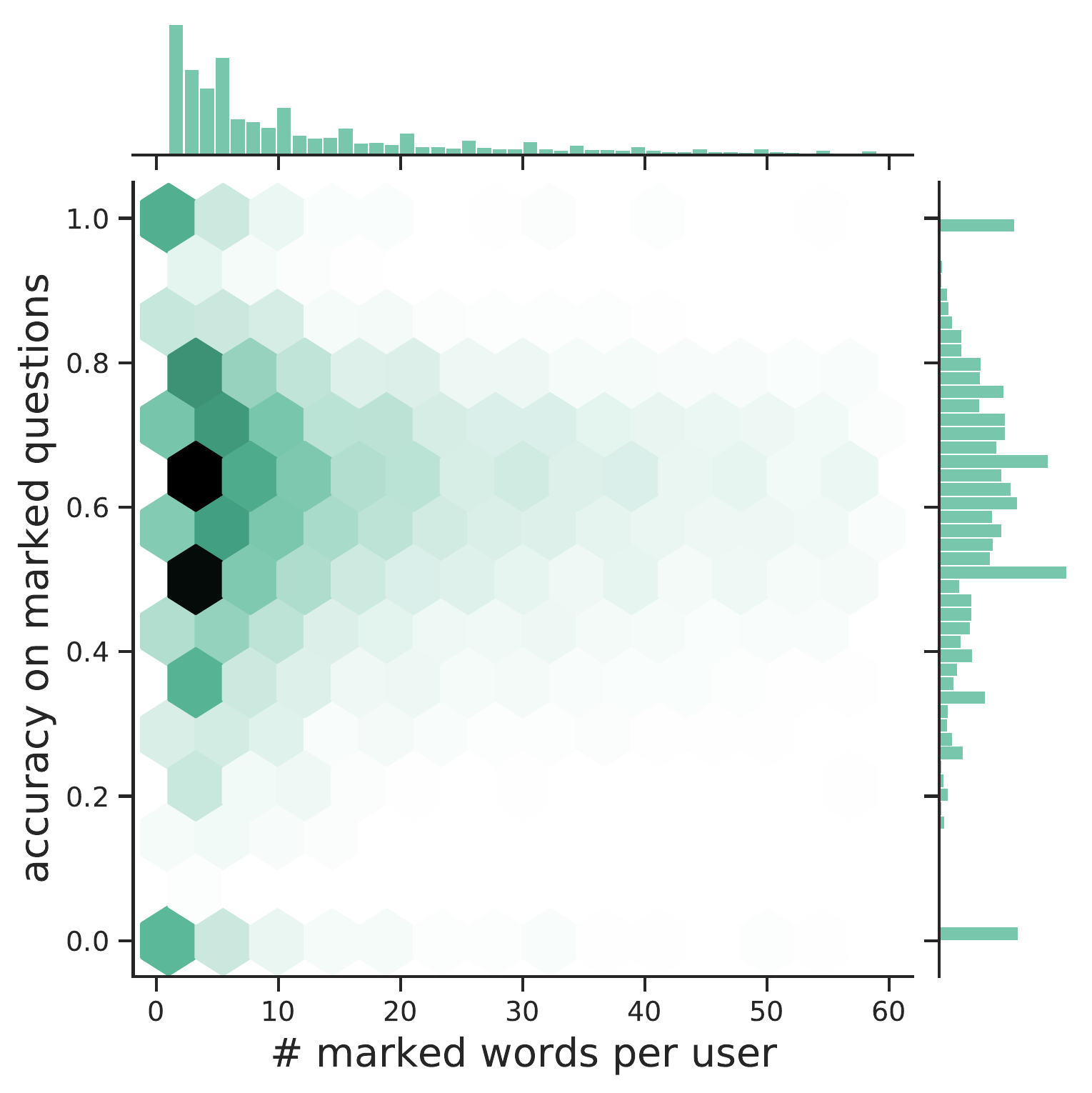}
\caption{Joint plot on the marginal distributions (hexbin plot version). The y-axis accuracy is calculated only on the questions that the marked words come from.}
\label{fig:hexbin}
\end{figure*}

\begin{figure*}[t]
\centering
\includegraphics[width=0.49\textwidth]{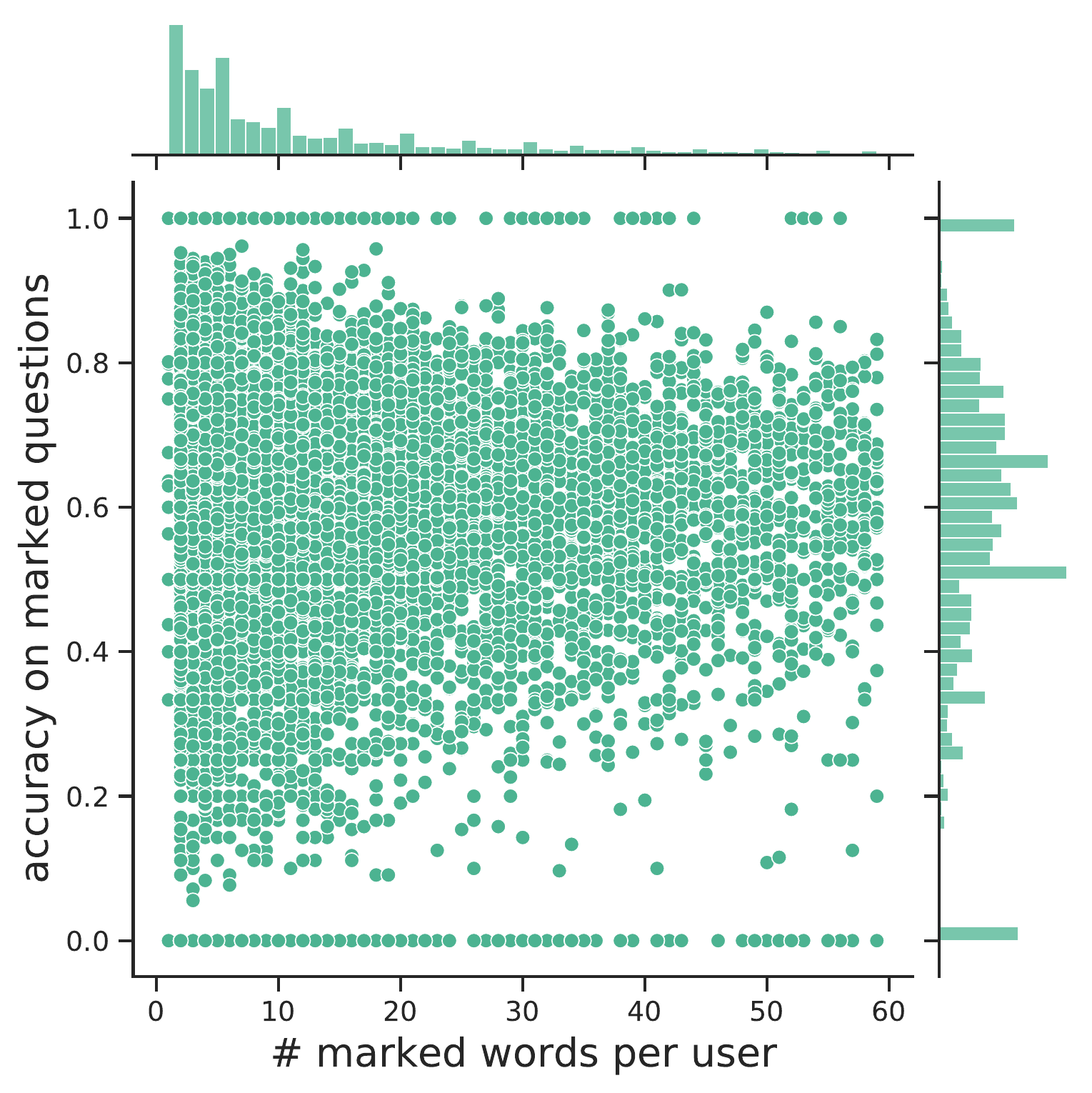}
\caption{Joint plot on the marginal distributions (scatter plot version). The y-axis accuracy is calculated only on the questions that the marked words come from.}
\label{fig:scatter}
\end{figure*}

\begin{figure*}[t]
\centering
\includegraphics[width=0.49\textwidth]{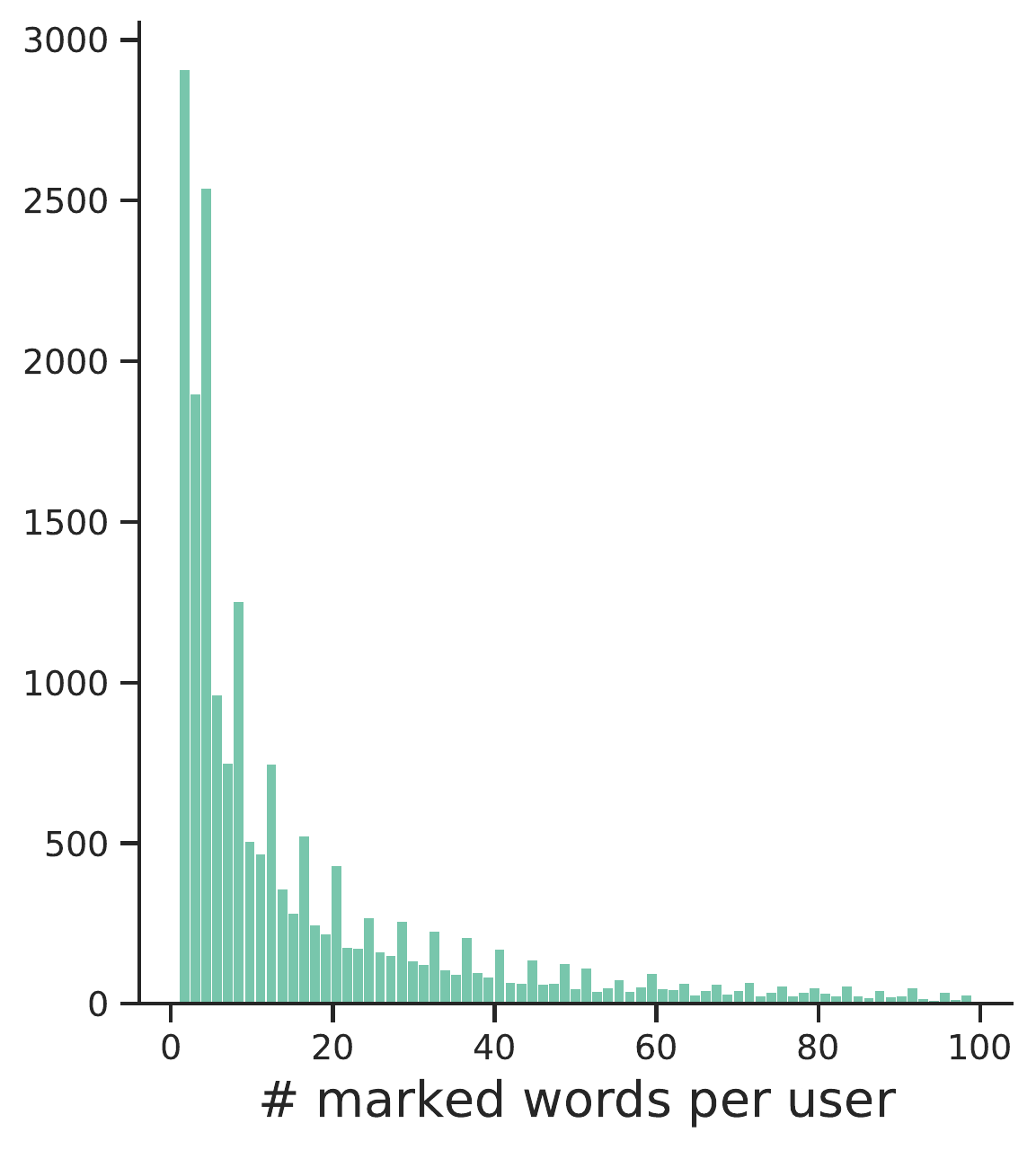}
\caption{Distribution of marked words per users.}
\label{fig:marked_words_per_users}
\end{figure*}

\begin{figure*}[t]
\centering
\includegraphics[width=0.49\textwidth]{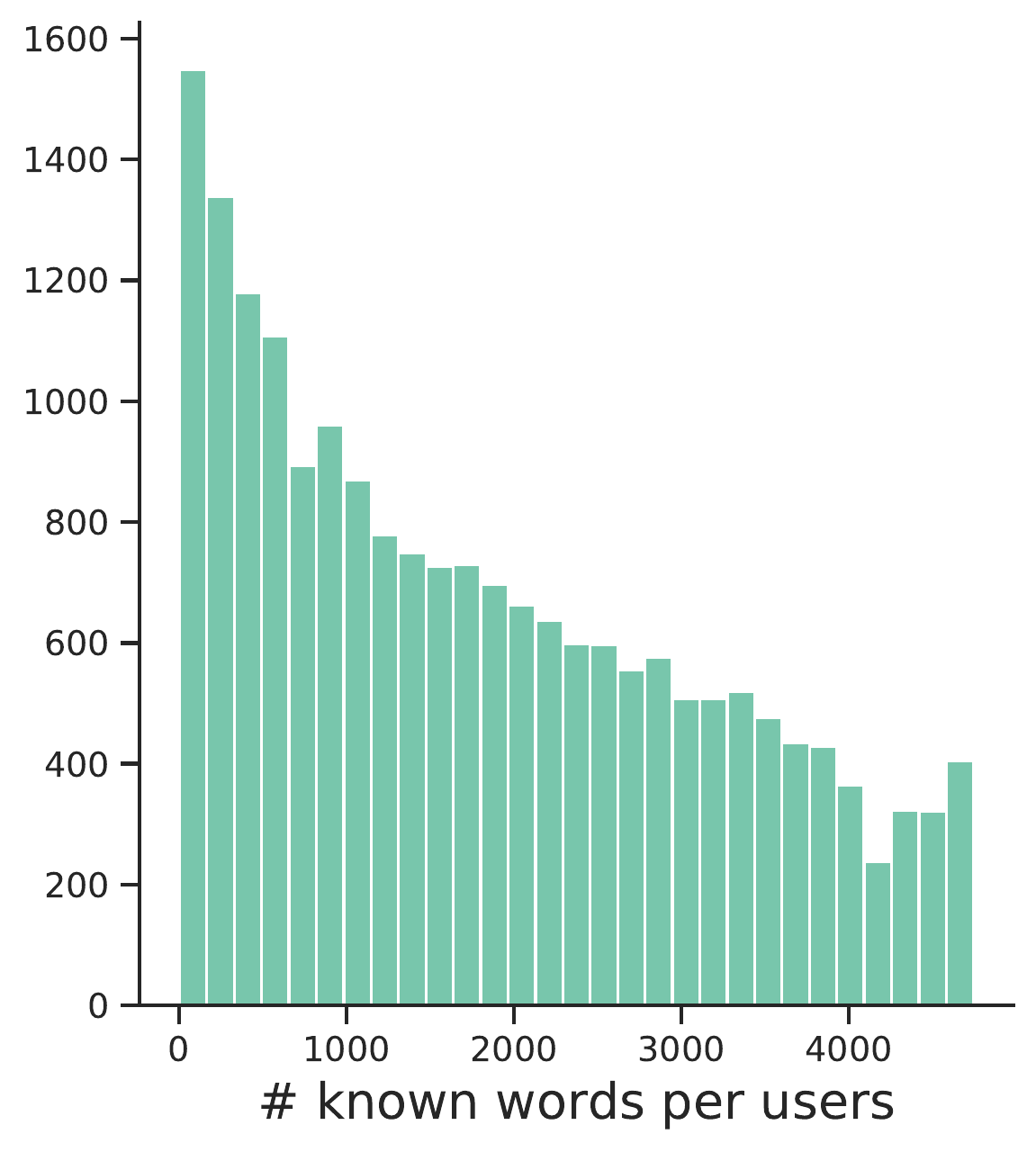}
\caption{Distribution of known words per users.}
\label{fig:known_words_per_users}
\end{figure*}

\begin{figure*}[t]
\centering
\includegraphics[width=0.49\textwidth]{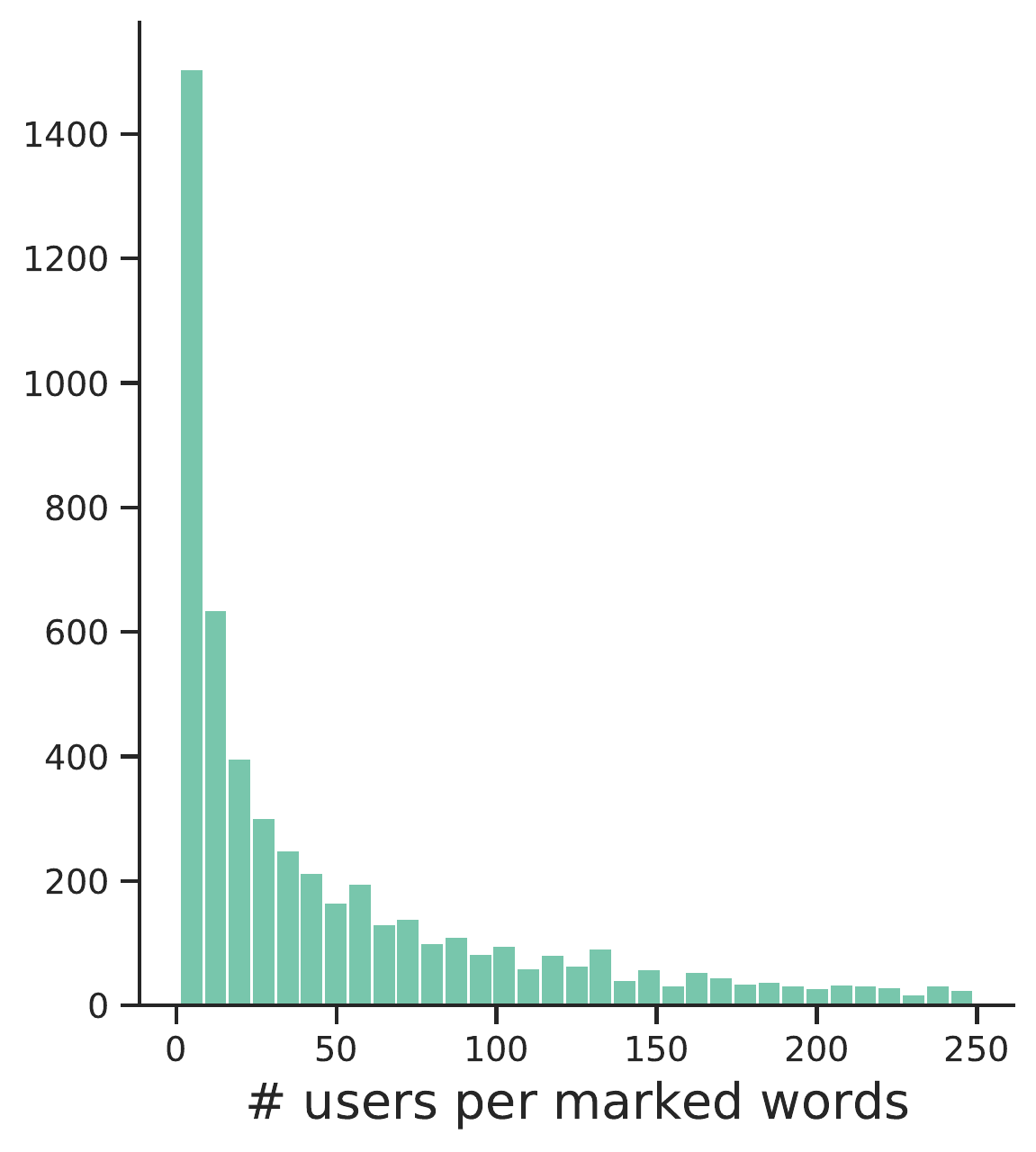}
\caption{Distribution of users per marked word.}
\label{fig:users_per_marked_words}
\end{figure*}

\begin{figure*}[t]
\centering
\includegraphics[width=0.49\textwidth]{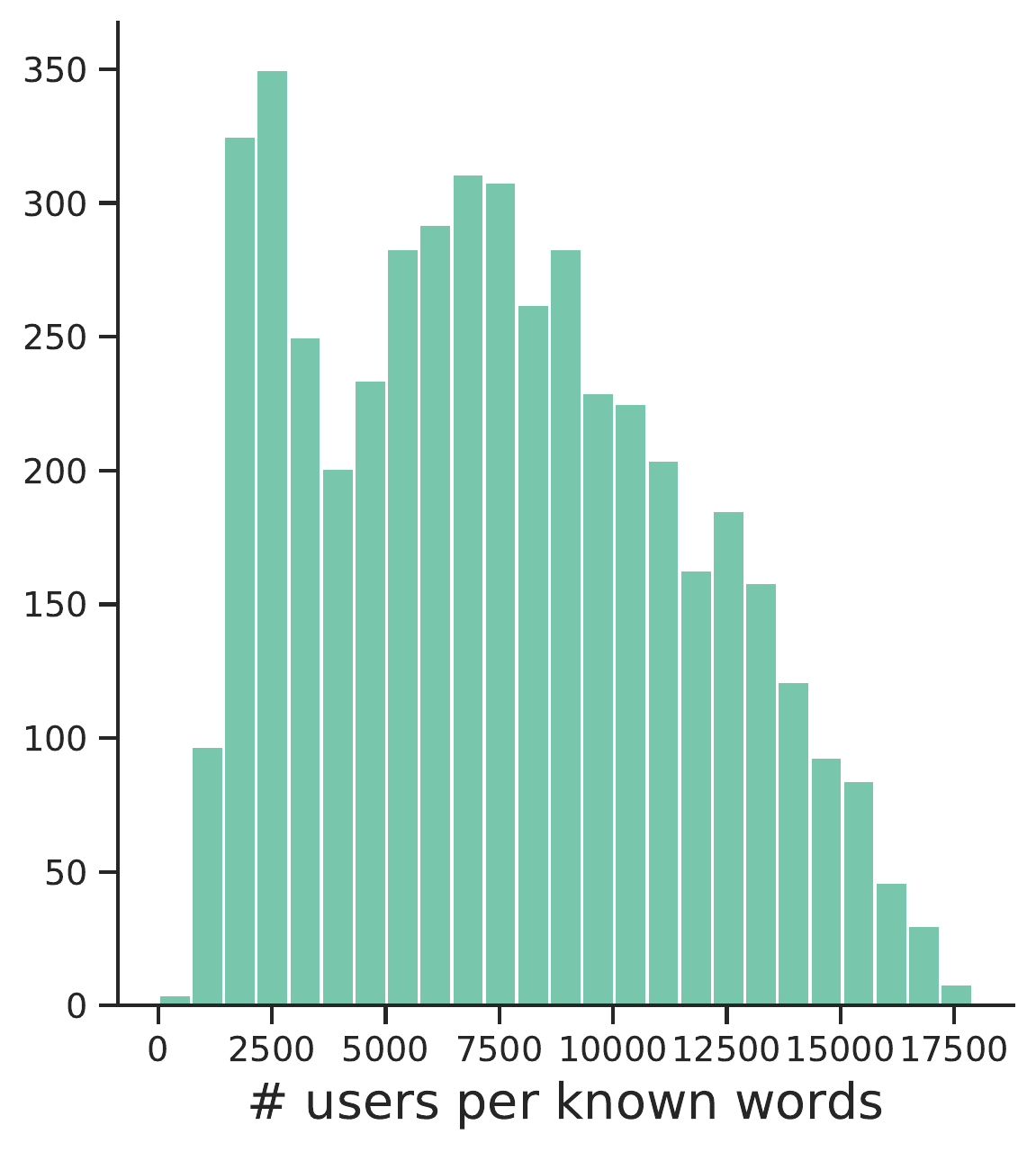}
\caption{Distribution of users per known word.}
\label{fig:users_per_known_words}
\end{figure*}

Figures~\ref{fig:hexbin} and~\ref{fig:scatter} show different versions of Figure~\ref{fig:kde}. Figures~\ref{fig:marked_words_per_users},~\ref{fig:known_words_per_users},~\ref{fig:users_per_marked_words},~\ref{fig:users_per_known_words} show the distribution plots related to the distributional statistics in Table~\ref{tab:statistics}.

\end{document}